\documentclass[runningheads]{llncs}
\usepackage{graphicx}
\usepackage{booktabs,multirow}
\usepackage{amsfonts,amssymb,amsmath}
\usepackage{float}
\usepackage{comment}
\usepackage{bbding}

\usepackage[pagebackref=true,breaklinks=true,colorlinks=true,bookmarks=false,citecolor=blue,urlcolor=black,linkcolor=blue]{hyperref}

\begin{document}
\title{Trans-SVNet: Accurate Phase Recognition from Surgical Videos via Hybrid Embedding Aggregation Transformer}
%
%
\author{Xiaojie Gao\inst{1}\and
	Yueming Jin  \inst{1} \and
	Yonghao Long\inst{1} \and
	Qi Dou\inst{1,2}\Envelope \and \\
	Pheng-Ann Heng\inst{1,2}}
%
\authorrunning{X. Gao et al.}
\institute{Department of Computer Science and Engineering,\\
	The Chinese University of Hong Kong, Hong Kong, China \\ \and
	T Stone Robotics Institute, CUHK, Hong Kong, China \\
	\email{qdou@cse.cuhk.edu.hk}
	}
	
\titlerunning{Accurate Surgical Phase Recognition via Transformer}
%

%
\maketitle              
\begin{abstract}
Real-time surgical phase recognition is a fundamental task in modern operating rooms. 
Previous works tackle this task relying on architectures arranged in spatio-temporal order, however, the supportive benefits of intermediate spatial features are not considered. 
In this paper, we introduce, for the first time in surgical workflow analysis, Transformer to reconsider the ignored complementary effects of spatial and temporal features for accurate surgical phase recognition. 
Our hybrid embedding aggregation Transformer fuses cleverly designed spatial and temporal embeddings by allowing for active queries based on spatial information from temporal embedding sequences. 
More importantly, our framework processes the hybrid embeddings in parallel to achieve a high inference speed. 
Our method is thoroughly validated on two large surgical video datasets, i.e., Cholec80 and M2CAI16 Challenge datasets, and outperforms the state-of-the-art approaches at a processing speed of 91 fps.

\end{abstract}
\section{Introduction}

With the developments of intelligent context-aware systems (CAS), the safety and quality of modern operating rooms have significantly been improved~\cite{maier2017surgical}.
One underlying task of CAS is surgical phase recognition, which facilitates surgery monitoring~\cite{bricon2007context}, surgical protocol extraction~\cite{zisimopoulos2018deepphase}, and decision support~\cite{padoy2019machine}.
However, purely vision-based recognition is quite tricky due to similar inter-class appearance and scene blur of recorded videos~\cite{jin2017sv,padoy2019machine}. 
Essentially, online recognition is even more challenging because future information is not allowed to assist current decision-making~\cite{yi2019hard}.
Moreover, processing high-dimensional video data is still time-consuming, given the real-time application requirement.

\begin{figure}[t]
	\centering
	\includegraphics[width=\textwidth]{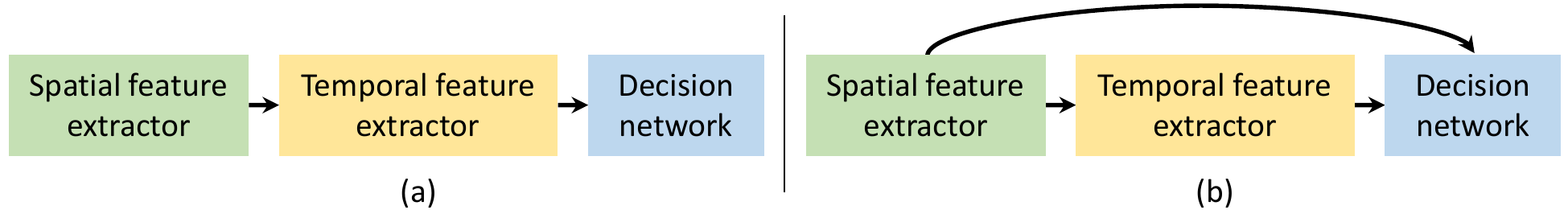}
	\caption{(a) Previous methods extract spatio-temporal features successively for surgical phase recognition; (b) We propose to reuse extracted spatial features together with temporal features to achieve more accurate recognition.}
	\label{fig:overview}
\end{figure}

Temporal information has been verified as a vital clue for various surgical video analysis tasks, such as robotic gesture recognition~\cite{funke2019using,gao2020automatic}, surgical instrument segmentation~\cite{jin2019incorporating,zhao2020learning}.
Initial methods for surgical workflow recognition, utilized statistical models, such as conditional random field~\cite{quellec2014real,charriere2017real} and hidden Markov models (HMMs)~\cite{padoy2008line,dergachyova2016automatic,twinanda2016endonet}. 
Nevertheless, temporal relations among surgical frames are highly complicated, and these methods show limited representation capacities with pre-defined dependencies~\cite{jin2017sv}.
Therefore, long short-term memory (LSTM)~\cite{hochreiter1997long} network was combined with ResNet~\cite{he2016deep} in SV-RCNet~\cite{jin2017sv} to model spatio-temporal dependences of video frames in an end-to-end fashion. 
Yi et al.~\cite{yi2019hard} suggested an Online Hard Frame Mapper (OHFM) based on ResNet and LSTM to focus on the pre-detected rigid frames. 
Gao et al.~\cite{gao2020automatic} devised a tree search algorithm to consider future information from LSTM for surgical gesture recognition. 
With additional tool presence labels, multi-task learning methods are proposed to boost phase recognition performance.
Twinanda~\cite{twinanda2017vision} replaced the HMM of EndoNet~\cite{twinanda2016endonet} with LSTM to enhance its power of modeling temporal relations.
MTRCNet-CL~\cite{jin2020multi}, the best multi-task framework, employed a correlation loss to strengthen the synergy of tool and phase predictions.
To overcome limited temporal memories of LSTMs, Convolutional Neural Networks (CNN) are leveraged to extract temporal features. 
Funke et al.~\cite{funke2019using} used 3D CNN to learn spatial and temporal features jointly for surgical gesture recognition.
Zhang et al.~\cite{zhang2020symmetric} devised a Temporal Convolutional Networks (TCN)~\cite{lea2016temporal,lea2017temporal} bridged with a self-attention module for offline surgical video analysis. 
Czempiel et al.~\cite{czempiel2020tecno} designed an online multi-stage TCN~\cite{farha2019ms} called TeCNO to explore long-term temporal relations in pre-computed spatial features. 
TMRNet~\cite{jin2021temporal}, a concurrent work, integrated multi-scale LSTM outputs via non-local operations.
However, these methods process spatial and temporal features successively, as shown in \autoref{fig:overview} (a), which leads to losses of critical visual attributes.


Transformer~\cite{vaswani2017attention} allows concurrently relating entries inside a sequence at different positions rather than in recurrent computing styles, which facilitates the preservation of essential features in overlong sequences. 
Therefore, it can enable the discovery of long-term clues for accurate phase recognition in surgical videos whose average duration spans minutes or hours. 
Moreover, thanks to its parallel computing fashion, high speed in both training and inference stages is realized. 
Besides strong capacity in sequence learning, Transformer also demonstrates outstanding ability in visual feature representation~\cite{han2020survey,khan2021transformers}.
Recently, Transformer was employed to fuse multi-view elements in point clouds and illustrated excellent outcomes~\cite{wang2019deep}, which implies its potential to promote the synergy of spatial and temporal features in surgical videos.

In this paper, we propose a novel method, named Trans-SVNet, for accurate phase recognition from \textbf{s}urgical \textbf{v}ideos via Hybrid Embedding Aggregation \textbf{Trans}former. 
As shown in \autoref{fig:overview} (b), we reconsider the spatial features as one of our hybrid embeddings to supply missing appearance details during temporal feature extracting.
Specifically, we employ ResNet and TCN to generate spatial and temporal embeddings, respectively, where representations with the same semantic labels cluster in the embedding space. 
Then, we introduce Transformer, for the first time, to aggregate the hybrid embeddings for accurate surgical phase recognition by using spatial embeddings to attend supporting information from temporal embedding sequences. 
More importantly, our framework is parameter-efficient and shows extraordinary potential for real-time applications.
We extensively evaluate our Trans-SVNet on two large public\footnote{http://camma.u-strasbg.fr/datasets} surgical video datasets. 
Our approach outperforms all the compared methods and achieves a real-time processing speed of 91 fps.

\section{Method}

\begin{figure}[t]	
	\centering
	\includegraphics[width=\textwidth]{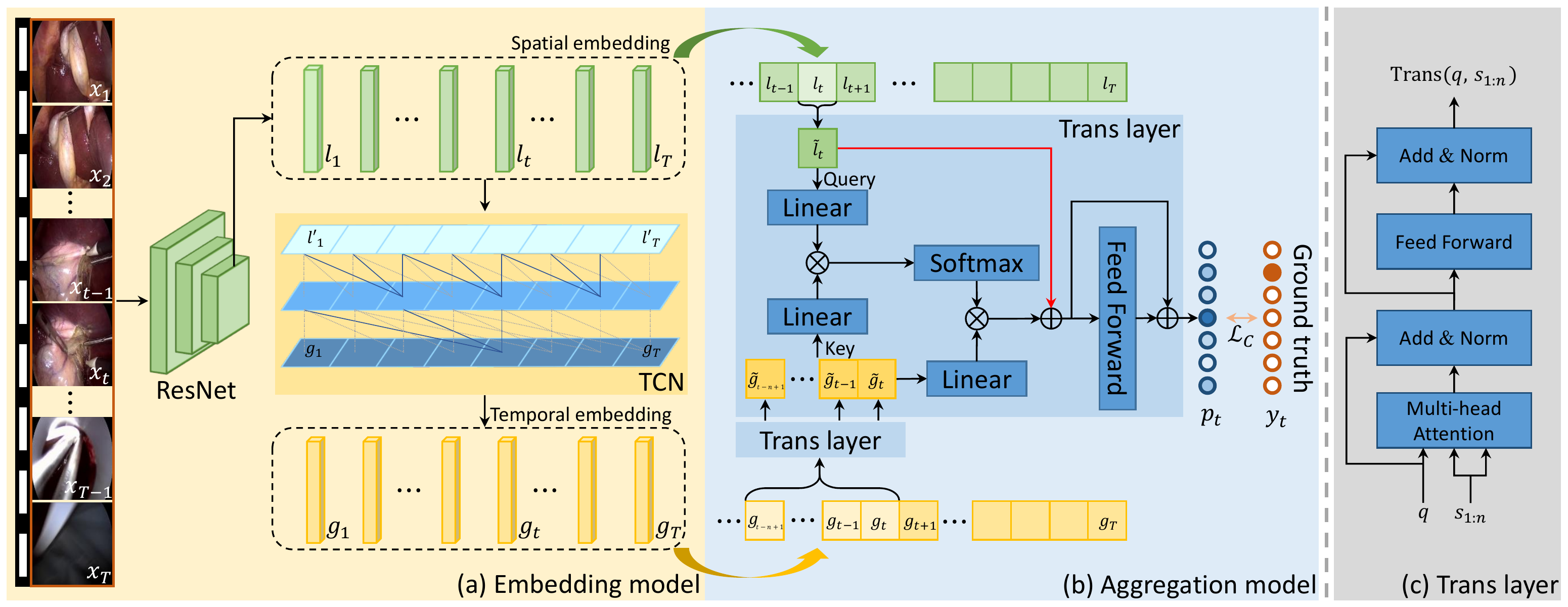}
	\caption{Overview of our proposed Trans-SVNet for surgical phase recognition. 
	(a) Extracted spatial embeddings enable the generation of temporal embeddings, (b) and are fused with the temporal information for refined phase predictions in our aggregation model, with the architecture of (c) Transformer layer presented in detail.
	}
    \label{fig:framework}
\end{figure}

\autoref{fig:framework} presents an overview of our proposed Trans-SVNet, composed of embedding and aggregation models. 
Our embedding model first represents surgical video frames with spatial embeddings $l$ and temporal embeddings $g$. The aggregation model fuses the hybrid embeddings by querying $l$ from $g$ to explore their synergy for accurate phase recognition.

\subsection{Transformer Layer}

Rather than only employed for temporal feature extraction, spatial features are reused to discover necessary information for phase recognition via our introduced Transformer. 
As depicted in~\autoref{fig:framework} (c), a Transformer layer, composed of a multi-head attention layer and a feed-forward layer, fuses a query $q$ with a temporal sequence $s_{1:n}=[s_1, \dots, s_{n-1}, s_n]$. 
Each head computes the attention of $q$ with $s_{1:n}$ as key and value:
\begin{equation}
	\begin{aligned}
		\mathrm{Attn}(q,s_{1:n})=\mathrm{softmax}(\frac{W_qq (W_ks_{1:n})^\mathrm{T}}{\sqrt{d_k}})W_vs_{1:n},
	\end{aligned}
\end{equation}
where $W$ are linear mapping matrices and $d_k$ is the dimension of $q$ after linear transformation. 
The outputs of all heads are concatenated and projected to enable the residual connection~\cite{he2016deep} with $q$ followed by a layer normalization~\cite{ba2016layer}. 
Since each attention head owns different learnable parameters, they concentrate on respective features of interest and jointly represent crucial features. 
We find it necessary to utilize multiple heads rather than a single head to produce a much faster convergence speed. 
The feed-forward layer is made up of two fully connected layers connected with a ReLU activation. 
The residual connection and layer normalization are applied in a similar way as the multi-head attention layer.
Finally, the output of the Transformer layer is denoted as $\mathrm{Trans}(q,s_{1:n})$, which contains synthesized information of $q$ and $s_{1:n}$.

\subsection{Video Embedding Extraction}

Given the discrete and sequential nature of video frames, we suggest two kinds of embeddings to represent their spatial and temporal information, which extends the spirit of word embeddings~\cite{mikolov2013distributed} to surgical video analysis. 
Let $x_t\in\mathbb{R}^{H\times W\times C}$ and $y_t\in\mathbb{R}^N$ denote the $t$-th frame of a surgical video with $T$ frames in total and the corresponding one-hot phase label, respectively. 
We first employ a very deep ResNet50~\cite{he2016deep} to extract discriminative spatial embeddings, which is realized by training a frame-wise classifier using the cross-entropy loss. 
Note that we only utilize phase labels because additional annotations like tool presence labels are not widely available, and single-task methods are more practical in real-world applications.
Then, outputs of the average pooling layer of ResNet50 are made as our spatial embeddings, i.e., $l_t\in \mathbb{R}^{2048}$, and high-dimensional video data are converted into low-dimensional embeddings.

To save memory and time, temporal embeddings are directly extracted from the spatial embeddings generated by the trained and fixed ResNet50. 
We first adjust the dimension of $l_t$ with a 1$\times$1 convolutional layer and generate $l'_t\in\mathbb{R}^{32}$.
Then, we exploit TCN to process the embedding sequence of a whole video without touching future information as illustrated in \autoref{fig:framework} (a). 
For easy comparison, we employ TeCNO~\cite{czempiel2020tecno}, a two-stage TCN model, to generate temporal embeddings using $l'_{1:T}$. 
Owing to multi-layer convolutions and dilated kernels, its temporal receptive field is increased to several minutes. 
Since $l_t$ is not updated, spatial embeddings of a whole video could be processed in a single forward computation, and the network converges quickly. 
Moreover, the outputs of the last stage of the TeCNO are used as our temporal embedding $g_{t}\in \mathbb{R}^{N}$.

\subsection{Hybrid Embedding Aggregation}

Our aggregation model, consisting of two Transformer layers, aims to output the refined prediction $p_t$ of frame $x_t$ by fusing the pre-computed hybrid video embeddings only available at time step $t$. 
The intuition is that a fixed-size representation encoded with spatio-temporal details is insufficient to express all critical features in both spatial and temporal dimensions, thus information loss is inevitably caused. 
Hence, we propose to look for supportive information based on a spatial embedding $l_t$ from an $n$-length temporal embedding sequence $g_{t-n+1:t}$ (see \autoref{sec:ablation} for ablation study), which allows for the rediscovery of missing yet crucial details during temporal feature extraction. 
In other words, our aggregation model learns a function $\mathbb{R}^{2048}\times\mathbb R^{n\times N}\rightarrow\mathbb{R}^{N}$.

Before synthesizing the two kinds of embeddings, they first conduct internal aggregation, respectively. 
On the one hand, dimension reduction is executed for the temporal embedding $l_t$ to generate $\tilde{l}_t\in\mathbb R^N$ by
\begin{equation}
	\begin{aligned}
		\tilde{l}_{t}=\mathrm{tanh}(W_ll_t),
	\end{aligned}
\end{equation}
where $W_l\in\mathbb R^{N\times2048}$ is a parameter matrix.
On the other hand, the temporal embedding sequence $g_{t-n+1:t}$ is processed by one of our Transformer layer to capture self-attention and an intermediate sequence $\tilde{g}_{t-n+1:t}\in\mathbb R^{n\times N}$ is produced.\footnote{Zero padding is applied if necessary.}
Specifically, each entry in $[g_{t-n+1},\dots,g_{t-1},g_t]$ attends all entries of the sequence, which is denoted as
\begin{equation}
	\begin{aligned}
        \tilde{g}_{i}=\mathrm{Trans}(g_{i}, g_{t-n+1:t}),~~i=t-n+1,\dots, t.
	\end{aligned}
\end{equation} 

Given self-aggregated embeddings $\tilde{l}$ and $\tilde{g}$, we employ the other Transformer layer to enable $\tilde l_t$ to query pivotal information from $\tilde g_{t-n+1:t}$ as key and value while fuse with the purified temporal features through residual additions (red arrow in \autoref{fig:framework} (b)).
Next, the output of the second Transformer layer is activated with the Softmax function to predict phase probability:
\begin{equation}
	\begin{aligned}
		p_{t}=\mathrm{Softmax}(\mathrm{Trans}(\tilde{l}_{t}, \tilde{g}_{t-n+1:t})).
	\end{aligned}
\end{equation}
Although the fused embeddings have a dimension of $N$, they still contain rich information for further processing. 
Lastly, our aggregation model is trained using the cross-entropy loss:
\begin{equation}
	\begin{aligned}
		\mathcal{L}_C=-\sum_{t=1}^{T}y_t \log(p_t).
	\end{aligned}
\end{equation}

\section{Experiments}

\subsubsection{Datasets.}
We extensively evaluate our Trans-SVNet on two challenging surgical video datasets of cholecystectomy procedures recorded at 25 fps, i.e., Cholec80~\cite{twinanda2016endonet} and M2CAI16 Challenge dataset~\cite{mi2cai}. 
Cholec80 includes 80 laparoscopic videos with 7 defined phases annotated by experienced surgeons.
Its frame resolution is either 1920$\times$1080 or 854$\times$480. 
This dataset also provides tool presence labels to allow for multi-task learning.
We follow the same evaluation procedure of previous works~\cite{twinanda2016endonet,jin2017sv,yi2019hard} by separating the dataset into the first 40 videos for training and the rest for testing.
The M2CAI16 dataset consists of 41 videos that are segmented into 8 phases by expert physicians.
Each frame has a resolution of 1920$\times$1080.
It is divided into 27 videos for training and 14 videos for testing, following the split of~\cite{twinanda2016single,jin2017sv,yi2019hard}. 
All videos are subsampled to 1 fps following previous works~\cite{twinanda2016endonet,jin2017sv}, and frames are resized into 250$\times$250.

\subsubsection{Evaluation Metrics.}
We employ four frequently-used metrics in surgical phase recognition for comprehensive comparisons.
These measurements are accuracy (AC), precision (PR), recall (RE), and Jaccard index (JA), which are also utilized in~\cite{jin2017sv,yi2019hard}.
The AC is calculated at the video level, defined as the percentage of frames correctly recognized in the entire video. 
Since the video classes are imbalanced, the PR, RE, and JA are first computed towards each phase and then averaged over all the phases. 
We also count the number of parameters to indicate the training and inference speed to a certain degree.

\begin{table}[t]
	\centering
	\caption{Phase recognition results (\%) of different methods on the Cholec80 and M2CAI16 datasets. The best results are marked in bold. Note that the * denotes methods based on multi-task learning that requires extra tool labels.}
	\label{tab:comparison}
	\resizebox{\textwidth}{!}{
		\begin{tabular}{l|cccc|cccc|c}
			\toprule
			\multirow{2}{*}{Method} & \multicolumn{4}{c|}{Cholec80}                                  &  \multicolumn{4}{c|}{M2CAI16} & \multirow{2}{*}{\#param}                                                                                                                                   \\ \cline{2-9} 
			& ~~Accuracy~~          & ~~Precision~~          & ~~~~Recall~~~~           & ~~Jaccard~~           & ~~Accuracy~~          & ~~Precision~~          & ~~~~Recall~~~~           & ~~Jaccard~~                                    \\ \midrule
			EndoNet*~\cite{twinanda2016endonet}                 & $81.7\pm4.2$                           & $73.7\pm16.1$                          & $79.6\pm7.9$                           & \rule[2pt]{0.35cm}{0.05em}   & \rule[2pt]{0.35cm}{0.05em}             & \rule[2pt]{0.35cm}{0.05em}            & \rule[2pt]{0.35cm}{0.05em}            & \rule[2pt]{0.35cm}{0.05em}            & 58.3M                                    \\
			EndoNet+LSTM*~\cite{twinanda2017vision}            & $88.6 \pm 9.6$                           & $84.4 \pm 7.9$                           & $84.7 \pm 7.9$                           & \rule[2pt]{0.35cm}{0.05em} & \rule[2pt]{0.35cm}{0.05em}             & \rule[2pt]{0.35cm}{0.05em}            & \rule[2pt]{0.35cm}{0.05em}           & \rule[2pt]{0.35cm}{0.05em}           &68.8M                                       \\
			MTRCNet-CL*~\cite{jin2020multi}              & $89.2 \pm 7.6$                           & $86.9 \pm 4.3$                           & $88.0 \pm 6.9$                           & \rule[2pt]{0.35cm}{0.05em}                                      & \rule[2pt]{0.35cm}{0.05em}             & \rule[2pt]{0.35cm}{0.05em}            & \rule[2pt]{0.35cm}{0.05em}            & \rule[2pt]{0.35cm}{0.05em}             &29.0M\\ \midrule
			PhaseNet~\cite{twinanda2016single,twinanda2016endonet}                & $78.8 \pm 4.7$                           & $71.3 \pm 15.6$                          & $76.6 \pm 16.6$                          & \rule[2pt]{0.35cm}{0.05em}      & $79.5 \pm 12.1$ & \rule[2pt]{0.35cm}{0.05em}            & \rule[2pt]{0.35cm}{0.05em}            & $64.1 \pm 10.3$      &58.3M                           \\
			SV-RCNet~\cite{jin2017sv}                  & $85.3 \pm 7.3$                           & $80.7 \pm 7.0$                           & $83.5 \pm 7.5$                           & \rule[2pt]{0.35cm}{0.05em}   & $81.7 \pm 8.1$  & $81.0 \pm 8.3$ & $81.6 \pm 7.2$ & $65.4 \pm 8.9$  &28.8M                                  \\
			OHFM~\cite{yi2019hard}                    & $87.3 \pm 5.7$                           & \rule[2pt]{0.35cm}{0.05em}                                     & \rule[2pt]{0.35cm}{0.05em}                                      & $67.0 \pm 13.3$    & $85.2 \pm 7.5$  & \rule[2pt]{0.35cm}{0.05em}            & \rule[2pt]{0.35cm}{0.05em}            & $68.8 \pm 10.5$      &47.1M                 \\ 
			TeCNO~\cite{czempiel2020tecno}                   & $88.6 \pm 7.8$                           & $86.5 \pm 7.0$                           & $87.6 \pm 6.7$                           & $75.1 \pm 6.9$                           &   $86.1 \pm 10.0$	            & $85.7 \pm 7.7$ 			    &$\mathbf{88.9\pm4.5}$   & $74.4 \pm 7.2$ 	&24.7M		  \\ 
			
			Trans-SVNet (ours)        & $\mathbf{90.3\pm7.1}$ & $\mathbf{90.7\pm5.0}$ & $\mathbf{88.8 \pm 7.4}$                           & $\mathbf{79.3\pm6.6}$                           &$\mathbf{87.2\pm9.3}$	   & $\mathbf{88.0\pm6.7}$   & $87.5 \pm 5.5$	 &$\mathbf{74.7 \pm 7.7}$	&24.7M \\
			\bottomrule
		\end{tabular}
	}

\end{table}

\subsubsection{Implementation Details.} 
Our embedding and aggregation models are trained one after the other on PyTorch using an NVIDIA GeForce RTX 2080 Ti GPU. 
We initialize the parameters of the ResNet from a pre-trained model on the ImageNet~\cite{he2016deep}. 
It employs an SGD optimizer with a momentum of 0.9 and a learning rate of 5e-4 except for its fully connected layers with 5e-5. 
Its batch size is set to 100, and data augmentation is applied, including 224$\times$224 cropping, random mirroring, and color jittering. 
We re-implement TeCNO~\cite{czempiel2020tecno} based on their released code with only phase labels and directly make outputs of its second stage as our temporal embeddings. 
We report the re-implemented results of TeCNO, and this well-trained model directly generates our temporal embeddings without further tuning.  
Our aggregation model is trained by Adam optimizer with a learning rate of 1e-3 and utilizes a batch size identical to the length of each video.
The number of attention heads is empirically set to 8, and the temporal sequence length $n$ is 30. $N$ is set to the dimension of the one-hot phase label. 
Our code is released at: \url{https://github.com/xjgaocs/Trans-SVNet}.

\subsubsection{Comparison with State-of-the-arts.} 
\autoref{tab:comparison} presents comparisons of our Trans-SVNet with seven existing methods without a post-processing strategy. 
Using extra tool presence annotations of the Cholec80 dataset, multi-task learning methods~\cite{twinanda2016endonet,twinanda2017vision,jin2020multi} generally achieve high performances, and MTRCNet-CL beats all single-task models except ours. 
As for methods using only phase labels, PhaseNet is far behind all other models due to its shallower network.
Thus the much deeper ResNet50 becomes a standard visual feature extractor since SV-RCNet~\cite{jin2017sv}.
As a multi-step learning framework like OHFM, our approach gains a significant improvement by 6\%-12\% in JA with a much simpler training procedure.
Compared to the state-of-the-art TeCNO with the same backbones, our Trans-SVNet gains a boost by 4\% in PR and JA on the larger Cholec80 dataset with a negligible increase in parameters ($\sim$30k). 
In a word, our Trans-SVNet outperforms all the seven compared methods, especially on the enormous Cholec80 dataset. 
Our method is observed to achieve a more remarkable improvement on the Cholec80 dataset than the M2CAI16 dataset. 
The underlying reason is that the M2CAI16 dataset is smaller and contains less challenging videos. 
The robustness of our method yields a better advantage on the more complicated Cholec80 dataset.
Thanks to the designed low-dimensional video embeddings, our model generates predictions at 91 fps with one GPU, which vastly exceeds the video recording speed. 

\begin{figure}[t]
	\centering
	\includegraphics[width=\textwidth]{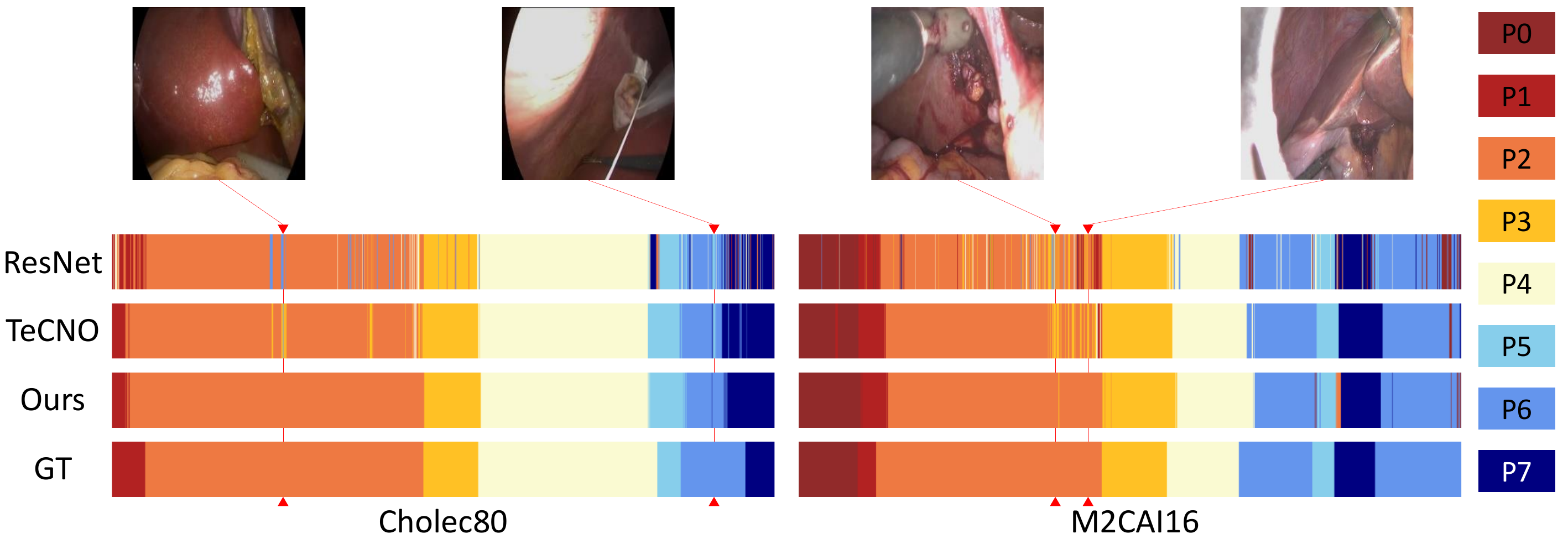}
	\caption{Color-coded ribbon illustration for two complete surgical videos. The time axes are scaled for better visualization.}
	\label{fig:colorbar}
\end{figure}

\subsubsection{Qualitative Comparison.}

In \autoref{fig:colorbar}, we show the color-coded ribbon of two complete laparoscopic videos from the two datasets. 
Due to the lack of temporal relations, ResNet suffers from noisy patterns and generates frequently jumped predictions.
TeCNO achieves smoother results by relating long-term temporal information in spatial embeddings generated by ResNet. However, its predictions for P2 in both videos still need to be improved. 
We also visualize some of the misclassified frames of TeCNO and find they are negatively influenced by excessive reflection, where bright but trivial parts might dominate the extracted spatial features, making it easy to miss pivotal information. 
Aggregating embeddings from ResNet and TeCNO elegantly, our Trans-SVNet contributes to more consistent and robust predictions of surgical phases, which highlights its promotion towards the synergy between the hybrid embeddings.

\subsubsection{Ablation Study.}
\label{sec:ablation}
We first analyze the effect of different length $n$ of our temporal embedding sequence on the Cholec80 dataset, and the results are reported in \autoref{tab:ablation_len}.
It is observed that our design of temporal sequence is undoubtedly necessary to gain a notable boost relative to not using temporal embeddings, i.e., $n=0$. 
We also notice that gradually increasing the temporal sequence length produces improvements towards all metrics, and our approach behaves almost equally well with the length $n\in[20, 40]$. 
The boost tends to be slower because adding $n$ by one increases the temporal sequence span by one second (only for $n>0$) and over-long sequences bring too much noise. 
Therefore, we choose $n=30$ as the length of our temporal embedding sequence.

\begin{table}[t]
	\centering
	\caption{Ablative testing results (\%) for increasing length of our temporal embedding sequence on the Cholec80 dataset.}
	\label{tab:ablation_len}
	\resizebox{0.4\textwidth}{!}{
		\begin{tabular}{c|cccc}
			\toprule
			Length ($n$) & ~Accuracy~ & ~Precision~ & ~~~~Recall~~~~ & ~~Jaccard~~ \\ \midrule
			0          & $82.1\pm7.8$                   & $78.0\pm6.4$                   & $78.5\pm10.8$                                         & $61.7\pm11.3$                  \\
			10         & $89.9 \pm 7.2$                   & $89.6 \pm 5.2$                    & $88.4 \pm 7.9$                                         & $78.4 \pm 6.6$                  \\
			20         & $90.2 \pm 7.1$                   & $90.2 \pm 5.1$                    & $\mathbf{88.8 \pm 7.7}$                                & $79.1 \pm 6.6$                  \\
			30         & $\mathbf{90.3 \pm 7.1}$          & $90.7 \pm 5.0$                    & $\mathbf{88.8 \pm 7.4}$                                & $\mathbf{79.3 \pm 6.6}$         \\
			40         & $\mathbf{90.3 \pm 7.0}$          & $\mathbf{90.8 \pm 4.9}$           & $88.5 \pm 7.2$                                         & $79.0 \pm 6.8$                  \\ \bottomrule
		\end{tabular}
	}
\end{table}

\begin{table}[t]
	\centering
	\caption{Phase recognition results (\%) of different architectures and their P-values in JA towards our proposed method on the Cholec80 dataset.}
	\label{tab:ablation}
	\resizebox{0.65\textwidth}{!}{
		\begin{tabular}{ccc|ccccc}
			\toprule
			\multicolumn{3}{c|}{Architecture}                                                          & ~~Accuracy~~ & ~~Precision~~ & ~~~~Recall~~~~ & ~~Jaccard~~ & P-values                                          \\ \midrule
			\multicolumn{1}{c|}{\multirow{3}{*}{PureNet}}     & \multicolumn{2}{c|}{ResNet}           & $82.1\pm7.8$                     & $78.0\pm6.4$                      & $78.5\pm10.8$                                          & $61.7\pm11.3$                   & 2e-8                                              \\
			\multicolumn{1}{c|}{}                             & \multicolumn{2}{c|}{TeCNO}            & $88.6\pm7.8$                     & $86.5\pm7.0$                      & $87.6\pm6.7$                                           & $75.1\pm6.9$                    & 2e-7                                              \\
			\multicolumn{1}{c|}{}                             & \multicolumn{2}{c|}{ResNet $cat$ TeCNO} & $87.9\pm7.5$                     & $86.6\pm5.9$                      & $85.3\pm8.2$                                           & $73.0\pm7.8$                    & 2e-8                                              \\ \midrule
			\multicolumn{1}{c|}{\multirow{5}{*}{Transformer}} & \multicolumn{1}{c|}{~~~Query~~~}   & Key    & Accuracy                         & Precision                         & Recall                                                 & Jaccard                         & P-values                                          \\ \cline{2-8} 
			\multicolumn{1}{c|}{}                             & \multicolumn{1}{c|}{$l_t$}  & $l_{t-n+1:t}$ & $81.9\pm9.2$                     & $78.0\pm12.5$                     & $78.3\pm12.8$                                          & $60.8\pm12.4$                   & 2e-8                                              \\
			\multicolumn{1}{c|}{}                             & \multicolumn{1}{c|}{$g_t$}   & $g_{t-n+1:t}$  & $89.1\pm7.8$                     & $87.6\pm6.3$                      & $87.7 \pm  6.9$                                        & $76.2\pm6.6$                    & 4e-7                                              \\
			\multicolumn{1}{c|}{}                             & \multicolumn{1}{c|}{$g_t$}   & $l_{t-n+1:t}$ & $89.2\pm7.5$                     & $87.7\pm6.7$                      & $87.7\pm7.0 $                                          & $76.1\pm7.0$                    & 3e-7                                              \\
			\multicolumn{1}{c|}{}                             & \multicolumn{1}{c|}{$l_t$}  & $g_{t-n+1:t}$  & $\mathbf{90.3\pm7.1}$            & $\mathbf{90.7\pm5.0}$             & $\mathbf{88.8\pm7.4}$                                  & $\mathbf{79.3\pm6.6}$           & \rule[2pt]{0.35cm}{0.05em} \\ \bottomrule
		\end{tabular}
	}
\end{table}

\autoref{tab:ablation} lists the results of different network structures, i.e., letting embeddings from ResNet and TeCNO be query or key in every possible combination, to identify which one makes the best use of information.
We first show baseline methods without Transformer denoted as PureNet.
ResNet $cat$ TeCNO employs a superficial linear layer to process concatenated $l_t$ and $g_t$, whose performance unsurprisingly falls between ResNet and TeCNO. 
As for Transformer-based networks, there are no advancements to use spatial embedding $l_t$ to query $l_{t-n+1:t}$. 
The reason is that spatial embeddings cannot indicate their orders in videos and bring ambiguity in the aggregation stage. 
Better performances are achieved than PureNet by letting TeCNO embeddings $g$ with sequential information be either query or key, which justifies Transformer rediscovers necessary details neglected by temporal extractors.
Our Trans-SVNet uses $l_t$ to query $g_{t-n+1:t}$ and generates the best outcomes with a clear margin, which confirms the effectiveness of our proposed architecture. 
We also calculate P-values in JA using Wilcoxon signed-rank test for compared settings towards our Trans-SVNet.
It is found that P-values are substantially less than 0.05 in all cases, which indicates that our model learns a formerly non-existent but effective policy.

\section{Conclusion}
We propose a novel framework to fuse different embeddings based on Transformer for accurate real-time surgical phase recognition.
Our novel aggregation style allows the retrieval of missing but critical information with rarely additional cost. 
Extensive experimental results demonstrate that our method consistently outperforms the state-of-the-art models while maintains a breakneck processing speed.
The excellent performance and parameter efficiency of our method justify its promising applications in real operating rooms.

\subsubsection{Acknowledgements.}
This work was supported by Hong Kong RGC TRS Project T42-409/18-R, and National Natural Science Foundation of China with Project No. U1813204.



\bibliographystyle{splncs04}
\bibliography{ref}

\begin{thebibliography}{10}
\providecommand{\url}[1]{\texttt{#1}}
\providecommand{\urlprefix}{URL }
\providecommand{\doi}[1]{https://doi.org/#1}

\bibitem{ba2016layer}
Ba, J.L., Kiros, J.R., Hinton, G.E.: Layer normalization. arXiv preprint
  arXiv:1607.06450  (2016)

\bibitem{bricon2007context}
Bricon-Souf, N., Newman, C.R.: {Context awareness in health care: A review}.
  International Journal of Medical Informatics  \textbf{76}(1),  2--12 (2007)

\bibitem{charriere2017real}
Charri{\`e}re, K., Quellec, G., Lamard, M., Martiano, D., Cazuguel, G.,
  Coatrieux, G., Cochener, B.: Real-time analysis of cataract surgery videos
  using statistical models. Multimedia Tools and Applications  \textbf{76}(21),
   22473--22491 (2017)

\bibitem{czempiel2020tecno}
Czempiel, T., Paschali, M., Keicher, M., Simson, W., Feussner, H., Kim, S.T.,
  Navab, N.: Tecno: Surgical phase recognition with multi-stage temporal
  convolutional networks. In: International Conference on Medical Image
  Computing and Computer-Assisted Intervention (2020)

\bibitem{dergachyova2016automatic}
Dergachyova, O., Bouget, D., Huaulm{\'e}, A., Morandi, X., Jannin, P.:
  Automatic data-driven real-time segmentation and recognition of surgical
  workflow. International Journal of Computer Assisted Radiology and Surgery
  (2016)

\bibitem{farha2019ms}
Farha, Y.A., Gall, J.: {MS-TCN: Multi-stage temporal convolutional network for
  action segmentation}. In: Proceedings of the IEEE/CVF Conference on Computer
  Vision and Pattern Recognition. pp. 3575--3584 (2019)

\bibitem{funke2019using}
Funke, I., Bodenstedt, S., Oehme, F., von Bechtolsheim, F., Weitz, J., Speidel,
  S.: {Using 3D convolutional neural networks to learn spatiotemporal features
  for automatic surgical gesture recognition in video}. In: International
  Conference on Medical Image Computing and Computer-Assisted Intervention
  (2019)

\bibitem{gao2020automatic}
Gao, X., Jin, Y., Dou, Q., Heng, P.A.: Automatic gesture recognition in
  robot-assisted surgery with reinforcement learning and tree search. In: IEEE
  International Conference on Robotics and Automation. pp. 8440--8446. IEEE
  (2020)

\bibitem{han2020survey}
Han, K., Wang, Y., Chen, H., Chen, X., Guo, J., Liu, Z., Tang, Y., Xiao, A.,
  Xu, C., Xu, Y., et~al.: A survey on visual transformer. arXiv preprint
  arXiv:2012.12556  (2020)

\bibitem{he2016deep}
He, K., Zhang, X., Ren, S., Sun, J.: Deep residual learning for image
  recognition. In: Proceedings of the IEEE Conference on Computer Vision and
  Pattern Recognition. pp. 770--778 (2016)

\bibitem{hochreiter1997long}
Hochreiter, S., Schmidhuber, J.: Long short-term memory. Neural Computation
  \textbf{9}(8),  1735--1780 (1997)

\bibitem{jin2019incorporating}
Jin, Y., Cheng, K., Dou, Q., Heng, P.A.: Incorporating temporal prior from
  motion flow for instrument segmentation in minimally invasive surgery video.
  In: International Conference on Medical Image Computing and Computer-Assisted
  Intervention (2019)

\bibitem{jin2017sv}
Jin, Y., Dou, Q., Chen, H., Yu, L., Qin, J., Fu, C.W., Heng, P.A.: {SV-RCNet:
  workflow recognition from surgical videos using recurrent convolutional
  network}. IEEE Transactions on Medical Imaging  \textbf{37}(5),  1114--1126
  (2018)

\bibitem{jin2020multi}
Jin, Y., Li, H., Dou, Q., Chen, H., Qin, J., Fu, C.W., Heng, P.A.: Multi-task
  recurrent convolutional network with correlation loss for surgical video
  analysis. Medical Image Analysis  \textbf{59},  101572 (2020)

\bibitem{jin2021temporal}
Jin, Y., Long, Y., Chen, C., Zhao, Z., Dou, Q., Heng, P.A.: Temporal memory
  relation network for workflow recognition from surgical video. IEEE
  Transactions on Medical Imaging  (2021)

\bibitem{khan2021transformers}
Khan, S., Naseer, M., Hayat, M., Zamir, S.W., Khan, F.S., Shah, M.:
  {Transformers in Vision: A Survey}. arXiv preprint arXiv:2101.01169  (2021)

\bibitem{lea2017temporal}
Lea, C., Flynn, M.D., Vidal, R., Reiter, A., Hager, G.D.: Temporal
  convolutional networks for action segmentation and detection. In: Proceedings
  of the IEEE Conference on Computer Vision and Pattern Recognition. pp.
  156--165 (2017)

\bibitem{lea2016temporal}
Lea, C., Vidal, R., Reiter, A., Hager, G.D.: Temporal convolutional networks: A
  unified approach to action segmentation. In: European Conference on Computer
  Vision. pp. 47--54. Springer (2016)

\bibitem{maier2017surgical}
Maier-Hein, L., Vedula, S.S., Speidel, S., Navab, N., Kikinis, R., Park, A.,
  Eisenmann, M., Feussner, H., Forestier, G., Giannarou, S., et~al.: Surgical
  data science for next-generation interventions. Nature Biomedical Engineering
   (2017)

\bibitem{mikolov2013distributed}
Mikolov, T., Sutskever, I., Chen, K., Corrado, G., Dean, J.: Distributed
  representations of words and phrases and their compositionality. In: Advances
  in Neural Information Processing Systems (2013)

\bibitem{padoy2019machine}
Padoy, N.: Machine and deep learning for workflow recognition during surgery.
  Minimally Invasive Therapy \& Allied Technologies  \textbf{28}(2),  82--90
  (2019)

\bibitem{padoy2008line}
Padoy, N., Blum, T., Feussner, H., Berger, M.O., Navab, N.: On-line recognition
  of surgical activity for monitoring in the operating room. In: Proceedings of
  the AAAI Conference on Artificial Intelligence. pp. 1718--1724 (2008)

\bibitem{quellec2014real}
Quellec, G., Lamard, M., Cochener, B., Cazuguel, G.: Real-time segmentation and
  recognition of surgical tasks in cataract surgery videos. IEEE Transactions
  on Medical Imaging  \textbf{33}(12),  2352--2360 (2014)

\bibitem{twinanda2016single}
Twinanda, A.P., Mutter, D., Marescaux, J., de~Mathelin, M., Padoy, N.:
  {Single-and multi-task architectures for surgical workflow challenge at M2CAI
  2016}. arXiv preprint arXiv:1610.08844  (2016)

\bibitem{mi2cai}
Twinanda, A.P., Shehata, S., Mutter, D., Marescaux, J., De~Mathelin, M., Padoy,
  N.: {MICCAI modeling and monitoring of computer assisted interventions
  challenge}. \url{http://camma.u-strasbg.fr/m2cai2016/}

\bibitem{twinanda2016endonet}
Twinanda, A.P., Shehata, S., Mutter, D., Marescaux, J., De~Mathelin, M., Padoy,
  N.: {Endonet: a deep architecture for recognition tasks on laparoscopic
  videos}. IEEE Transactions on Medical Imaging  \textbf{36}(1),  86--97 (2017)

\bibitem{twinanda2017vision}
Twinanda, A.P.: {Vision-based approaches for surgical activity recognition
  using laparoscopic and RBGD videos}. Ph.D. thesis, Strasbourg (2017)

\bibitem{vaswani2017attention}
Vaswani, A., Shazeer, N., Parmar, N., Uszkoreit, J., Jones, L., Gomez, A.N.,
  Kaiser, L., Polosukhin, I.: Attention is all you need. In: Advances in Neural
  Information Processing Systems. pp. 5998--6008 (2017)

\bibitem{wang2019deep}
Wang, Y., Solomon, J.M.: Deep closest point: Learning representations for point
  cloud registration. In: Proceedings of the IEEE/CVF International Conference
  on Computer Vision. pp. 3523--3532 (2019)

\bibitem{yi2019hard}
Yi, F., Jiang, T.: Hard frame detection and online mapping for surgical phase
  recognition. In: International Conference on Medical Image Computing and
  Computer-Assisted Intervention (2019)

\bibitem{zhang2020symmetric}
Zhang, J., Nie, Y., Lyu, Y., Li, H., Chang, J., Yang, X., Zhang, J.J.:
  Symmetric dilated convolution for surgical gesture recognition. In:
  International Conference on Medical Image Computing and Computer-Assisted
  Intervention (2020)

\bibitem{zhao2020learning}
Zhao, Z., Jin, Y., Gao, X., Dou, Q., Heng, P.A.: Learning motion flows for
  semi-supervised instrument segmentation from robotic surgical video. In:
  International Conference on Medical Image Computing and Computer-Assisted
  Intervention. pp. 679--689. Springer (2020)

\bibitem{zisimopoulos2018deepphase}
Zisimopoulos, O., Flouty, E., Luengo, I., Giataganas, P., Nehme, J., Chow, A.,
  Stoyanov, D.: {DeepPhase: surgical phase recognition in CATARACTS videos}.
  In: International Conference on Medical Image Computing and Computer-Assisted
  Intervention (2018)

\end{thebibliography}

\title{Supplementary Materials}
	%
	%
\author{}
\authorrunning{}
	%
\institute{}
\maketitle              

\renewcommand{\thefigure}{S\arabic{figure}}

\begin{figure}[ht]	
		\centering
		\includegraphics[width=\textwidth]{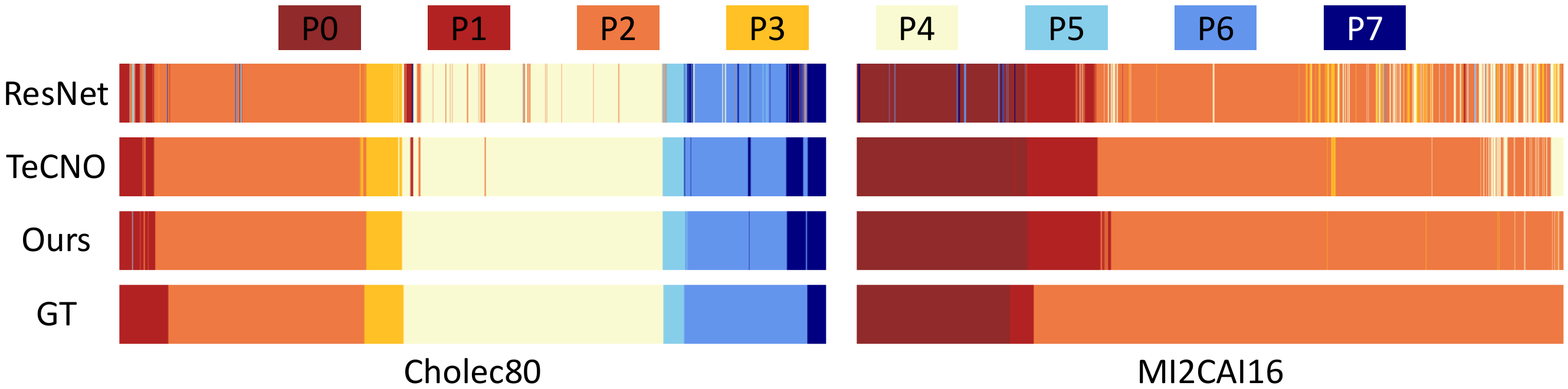}
		\caption{Additional results for qualitative comparisons on the two datasets.
		}
\end{figure}

\begin{figure}[ht]	
		\centering
		\includegraphics[width=\textwidth]{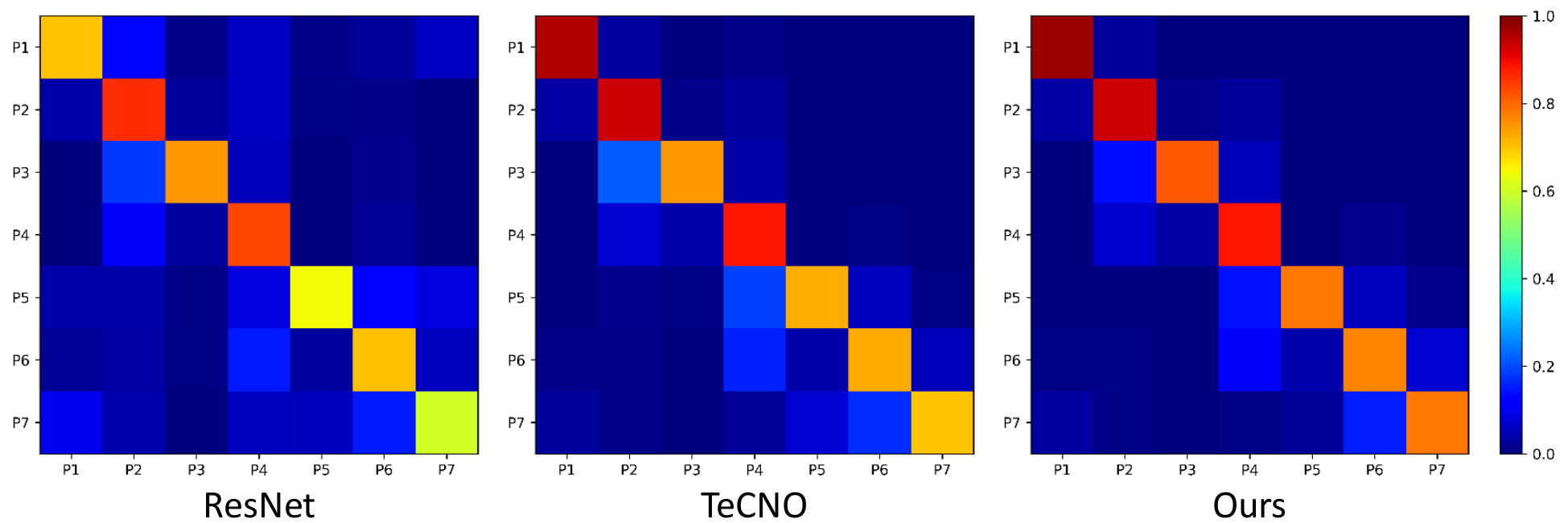}
		\caption{
		Confusion matrices visualized by color brightness on the Cholec80 dataset.
		}
\end{figure}

\begin{figure}[ht]	
	\centering
	\includegraphics[width=\textwidth]{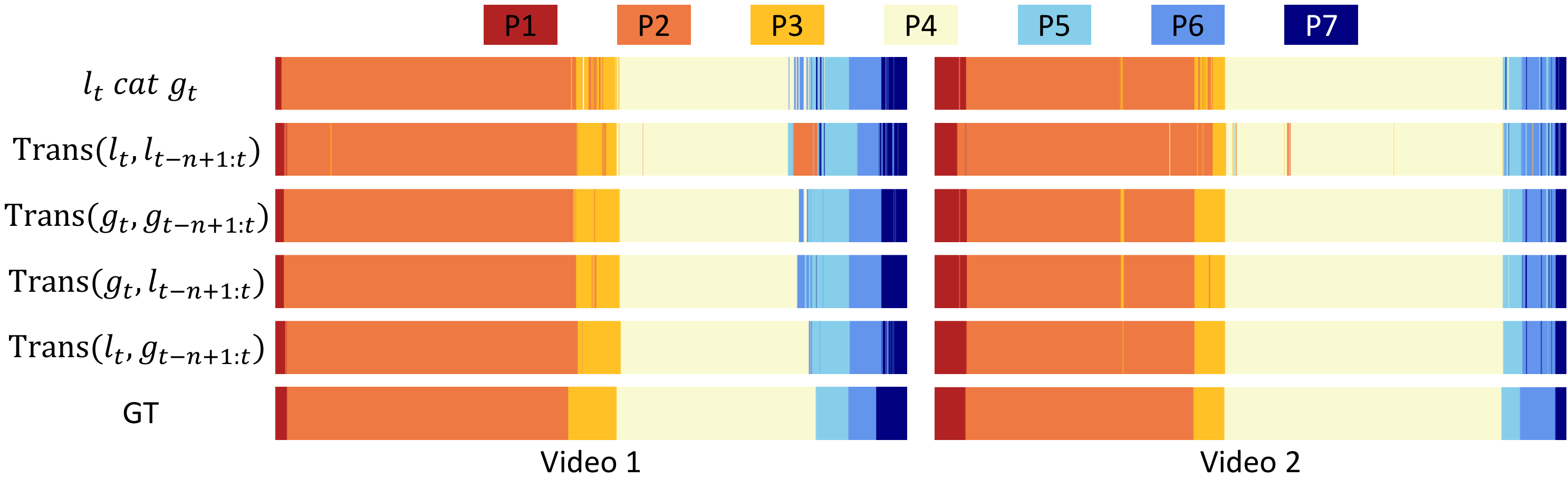}
	\caption{Visual results for ablation settings on the Cholec80 dataset.
	}
\end{figure}

\end{document}